 \documentclass[final,5p,times,twocolumn]{elsarticle}

\usepackage{amssymb}
\usepackage{lipsum}
\usepackage{lscape}
\usepackage{graphicx}%
\usepackage{multirow}%
\usepackage{amsmath,amssymb,amsfonts}%
\usepackage{amsfonts}
\usepackage{mathrsfs}%
\usepackage{textcomp}%
\usepackage{manyfoot}%
\usepackage{booktabs}%
\usepackage{algorithm}%
\usepackage{algorithmicx}%
\usepackage{algpseudocode}%
\usepackage{listings}%
\usepackage{url}
\usepackage{physics}
\usepackage{amsmath}
\usepackage{tikz}
\usepackage{mathdots}
\usepackage{yhmath}
\usepackage{cancel}
\usepackage{color}
\usepackage{siunitx}
\usepackage{array}
\usepackage{multirow}
\usepackage{amssymb}
\usepackage{gensymb}
\usepackage{tabularx}
\usepackage{extarrows}
\usepackage{booktabs}
\usetikzlibrary{fadings}
\usetikzlibrary{patterns}
\usetikzlibrary{shadows.blur}
\usetikzlibrary{shapes}
\usepackage{listings}
\usepackage{graphicx} 
\usepackage{fancybox}
\usepackage{adjustbox}
\usepackage{twoopt} 
\usepackage{subcaption}
\usepackage{amsfonts} 
\usepackage{amssymb}
\usepackage{booktabs, multirow} 
\usepackage{soul}
\usepackage{changepage,threeparttable} 
\usepackage{forest}
\usepackage{pdflscape}
\usepackage{adjustbox}
\usetikzlibrary{shadows, trees, positioning}
\usepackage{booktabs}
\usepackage{xltabular}
\usepackage{tabularray}
\usepackage{enumerate}
\usetikzlibrary{shapes,arrows,positioning}
\usepackage[many]{tcolorbox}
\usetikzlibrary{calc}
\tcbuselibrary{skins}
\usepackage{dashrule}
\usetikzlibrary{decorations.pathreplacing}
\usepackage{float}

\newcommandtwoopt{\insertboxtwo}[3][1.0][4.0cm]{\noindent\fbox{\begin{minipage}[c]{#1\textwidth}\parbox[c][#2]{#1\textwidth}{#3}\hfill\end{minipage}}}

\journal{Nuclear Physics B}

\newcommand{\firstletter}[2]{#1#2} 
\begin{document}

\begin{frontmatter}



\title{It’s All in the Way You Say It: The Role of Information Representation in LLM-Based Glycemic-Event Prediction}

\affiliation[unisa]{organization={Department of Information Engineering, Electrical Engineering, and Applied Mathematics (DIEM), University of Salerno},
city={Fisciano (SA)},
country={Italy}}
\affiliation[unina]{organization={Department of Electrical Engineering and Information Technology, University of Naples Federico II},
city={Naples},
country={Italy}}
\author[unisa]{Andrea Apicella}
\author[unina]{Pasquale Arpaia}

\author[unina]{Matteo Orefice}

\author[unina]{Andrea Pollastro}

\author[unina]{Roberto Prevete}

\begin{abstract}
Large Language Models (LLMs) are increasingly being investigated for physiological time-series prediction, yet their effectiveness may depend not only on the model itself, but also on how physiological information is represented and presented at inference time. This study investigates prompt-based general-purpose LLMs for postprandial hyperglycemia and hypoglycemia prediction in individuals with type 1 diabetes. Using the OhioT1DM dataset, we evaluate multiple open-weight LLMs under zero-shot and few-shot inference across prediction horizons of 30, 60, and 90 minutes. The analysis varies both the textual representation of the available physiological information and the amount of information exposed to the model, ranging from glucose observations alone to derived descriptors and additional contextual variables related to insulin, meals, carbohydrates, and physical activity. Performance is compared with conventional patient-specific supervised models and with Gluco-LLM, a language-model-based architecture explicitly adapted to glucose time-series forecasting. Results show a marked task-dependent behavior. Conventional supervised models achieve the strongest performance for hyperglycemia prediction, whereas the best observed prompt-based LLM configurations improve performance for hypoglycemia across all investigated horizons. The effectiveness of prompt-based inference is also strongly influenced by how physiological information is represented, while providing additional contextual information does not lead to a systematic improvement. Overall, these findings highlight physiological information representation as a central design factor in prompt-based LLM approaches to glycemic-event prediction.
\end{abstract}




\end{frontmatter}

\section{Introduction}
\label{sec:introduction}
{\def\thefootnote{}\footnotetext{This work has been submitted to a journal for peer review.}}\firstletter{T}{ype} 1 diabetes (T1D) is a chronic autoimmune condition characterized by the destruction of pancreatic $\beta$-cells, resulting in little or no endogenous insulin production \cite{atkinson2014type}. Consequently, blood glucose regulation relies on exogenous insulin administration and continuous monitoring of glucose levels. Inadequate glucose control may lead to hyperglycemia, i.e., excessively high blood glucose levels, or hypoglycemia, i.e., abnormally low glucose levels, both of which can have clinically relevant consequences. Continuous Glucose Monitoring (CGM) systems provide frequent measurements of interstitial glucose and therefore represent a key source of information for anticipating such adverse glycemic events.
Large Language Models (LLMs) have recently emerged as a promising alternative for addressing a wide range of medical prediction and decision-support tasks \cite{gu2021domain,alsentzer2019publicly,peng2019transfer,lee2020biobert}. Importantly, LLM-based and LLM-assisted approaches have attracted increasing attention for blood glucose forecasting in people with T1D,  
showing encouraging results in personalized prediction and in the integration of heterogeneous physiological information \cite{lara2025personalized,wolber2025multimodal}. This development is particularly relevant given the growing global burden of T1D \cite{ogle2025global} and the challenges that conventional Machine Learning (ML) approaches still face in capturing highly individual glucose dynamics and effectively exploiting heterogeneous physiological inputs \cite{nemat2024data,cinar2025review, annuzzi2023exploring,annuzzi2023impact}. 
In this work, we specifically focus on postprandial glycemic-event prediction, where the objective is to anticipate hyperglycemic and hypoglycemic events following meal intake.  The postprandial period represents a particularly challenging and clinically relevant setting, as glucose dynamics are strongly affected by the interplay between several factors, such as carbohydrate absorption, insulin administration, and individual physiological responses. Accurate prediction in this phase may therefore support earlier identification of potentially adverse glycemic excursions and enable more timely preventive interventions \cite{oviedo2019risk}.

Furthermore, general purpose LLMs may offer a practical advantage in terms of accessibility. Unlike conventional supervised models, they can be applied to a new task without task-specific parameter training and can expose their predictive capabilities through a natural language interface. Thus, this interaction paradigm can mitigate the technical barrier for non-specialist users, who would not be required to configure, train, and execute a dedicated ML pipeline. In the context of postprandial glycemic-event prediction, however, accessibility also depends on how physiological information can be communicated to the model. If clinically relevant descriptors can be expressed in natural language without a substantial loss in predictive performance, LLM-based systems could support the development of more intuitive user tools. 
Moreover, LLM-based inference also offers practical properties that are not captured by predictive performance alone. A general purpose LLM can address the task without patient-specific parameter training, potentially reducing the computational and technical effort required to train and maintain a dedicated predictive model. 
Unlike conventional ML algorithms, however, general purpose LLMs are not explicitly designed to process numerical physiological time series such as Continuous Glucose Monitoring (CGM) signals. Consequently, before they can be employed for glucose prediction, physiological observations must first be transformed into a textual representation that can be interpreted through natural language processing.

This transformation is not merely an implementation detail. The effectiveness of prompt-based processing depends on two complementary factors. The first concerns \emph{how} physiological observations are represented before being presented to the LLM. For example, the original CGM sequence may be provided directly, summarized through clinically meaningful descriptors, or verbalized as natural language statements. The second concerns \emph{which} physiological and clinical information is made available to the model. Besides the glucose measurements themselves, additional descriptors or contextual variables related to insulin administration, carbohydrate intake, and physical activity may also be incorporated. These two dimensions jointly determine the evidence available to the LLM and can substantially influence its prediction capabilities.

Although several recent studies have explored the use of LLMs for glucose prediction in T1DM and Type 2 Diabetes Mellitus (T2DM), existing approaches generally follow one of two directions. The first adapts pretrained LLM to time series forecasting through task-specific optimization \cite{li2025llm,gao2026llm,mahmoudi2026diabllm,lara2025personalized}. The second directly queries LLMs through proper prompts \cite{healey2025case,alredaini2025interpretable} and evaluates their predictions against conventional supervised learning approaches. While both directions have demonstrated encouraging results, they provide only limited understanding of a more fundamental question: \emph{how should physiological information be represented before being presented to a language model, and how much physiological information is actually required to support reliable clinical prediction?}

This work addresses these issues from a different perspective. Rather than proposing a new prediction architecture or further adapting an existing LLM, this work provides an empirical evaluation of how physiological information and its different representations can affect prompt-based processing. To this aim, we exploit the multimodal nature of the OhioT1DM dataset~\cite{marling2020ohiot1dm}, which provides CGM measurements together with insulin therapy data, meal and carbohydrate information, self-reported behavioral and contextual events, and wearable-derived physiological signals. In the present study, a selected subset of these modalities is used to define alternative information settings, ranging from CGM-derived information to configurations augmented with additional clinical and behavioral variables.

To provide a comprehensive assessment, prompt-based inference is evaluated together with two complementary prediction paradigms. The first consists of conventional supervised ML models trained specifically for postprandial glycemic-event prediction. The second is represented by Gluco-LLM \cite{li2025llm}, a language model explicitly adapted to physiological T1DM time series forecasting through task-specific architectural modifications. Comparing these three paradigms allows us to assess the potential of prompt-based processing and whether task-specific adaptation remains necessary for accurate glucose prediction.

The experimental evaluation is organized around three complementary research questions:

\begin{itemize}

\item \textbf{RQ1.} Can current general purpose LLMs provide predictive performance comparable to conventional supervised learning models for postprandial glycemic-event prediction?

\item \textbf{RQ2.} How do the representation of physiological information and the type of available physiological information in LLMs influence the capabilities of current general purpose LLMs?

\item \textbf{RQ3.} Can a non-adapted general purpose LLM achieve predictive performance comparable to an LLM explicitly adapted for physiological time series forecasting, such as Gluco-LLM?
\end{itemize}

To address RQ1, we evaluate general purpose LLMs with respect to conventional supervised learning, treating these approaches as alternative prediction paradigms rather than simply performance baselines. To address RQ2, we analyze three alternative representation strategies, namely \emph{Raw}, \emph{Structured}, and \emph{Narrative}. They represent three different and complementary ways to build the LLM input (prompt) from raw physiological data. \emph{Raw} preserves the original physiological data via textual serialization, while \emph{Structured} and \emph{Narrative} represent derived physiological descriptors in terms of \textit{key–value} pairs and \textit{natural language}, respectively. The latter two representations are provided in two different modalities corresponding to different sets of physiological information, namely  $CGM_{derived}$ and \emph{$CGM_{derived}^{context}$}: the former contains descriptors derived from the historical CGM signal, whereas the latter additionally incorporates selected contextual variables related to insulin administration, carbohydrate intake, and physical activity (see Sec. \ref{sec:method} for further details). Finally, to address RQ3 we investigate whether prompt-based processing with general purpose LLMs can approach the predictive capability of a language model specifically adapted to physiological time series  forecasting, thereby quantifying the benefits provided by task-specific architectural adaptation.

Our results show that the relative effectiveness of the considered approaches depends on the prediction task. Conventional supervised models achieve the strongest performance for postprandial hyperglycemia prediction, whereas general purpose LLMs are particularly competitive for hypoglycemia prediction. We further observe that LLM performance is sensitive to how physiological information is represented, with no representation proving uniformly optimal across models, tasks, and prompting conditions. When performance is summarized by selecting the best observed result across LLM families, \emph{Structured} inputs are most frequently associated with the strongest configurations. However, within matched model configurations, \emph{Raw} inputs become more competitive under few-shot inference, where a small number of labeled examples are provided in the prompt as in-context demonstrations without updating the model parameters~\cite{brown2020language}, while \emph{Narrative} representations remain preferable in selected settings. Finally, Gluco-LLM generally outperforms prompt-based general purpose LLMs in the regression experiments, although general purpose prompt-based inference remains competitive in selected conditions.

\section{Related Works}
\label{sec:related}

Glucose forecasting in T1D has been extensively investigated using ML approaches and evaluated across multiple prediction horizons~\cite{plis2014machine,martinsson2020blood,de2022glyfe,voegeli2025beyond}.
Cui et al.~\cite{cui2023jointly} addressed the joint prediction of postprandial hyperglycemia and hypoglycemia from CGM data, proposing a patient-specific LSTM model that uses historical glucose measurements to estimate the maximum and minimum glucose values within a given prediction horizon. Their formulation provides an explicit protocol for constructing postprandial prediction instances and defining hyperglycemic and hypoglycemic events under a 30-minute prediction horizon.

More recently, several studies have demonstrated that general purpose LLMs can be applied to time series forecasting by representing numerical observations as textual sequences. For instance, Gruver et al. \cite{gruver2023large} introduced LLMTime, showing how general purpose LLMs can perform zero-shot time series forecasting by representing numerical sequences as text and formulating forecasting as a next-token prediction task. Their work established a paradigm for treating numerical time series as textual sequences for prompt-based forecasting, providing a methodological foundation that can also be applied to physiological time series tasks.
Alredaini et al. \cite{alredaini2025interpretable} compared traditional ML, DL, and LLMs for multi-horizon glucose forecasting. The authors formulate the task as a natural language reasoning problem by encoding historical CGM measurements and patient-specific characteristics into textual prompts, allowing fine-tuned LLMs such as GPT-4.1 \cite{openai2025gpt41} and LLaMA \cite{grattafiori2024llama} to directly predict future glucose values, showing the feasibility of prompt-based formulations for physiological time series prediction.
Lara-Abelenda et al. \cite{lara2025personalized} investigated the use of Time-LLM for personalized glucose forecasting in people with T1D reprogramming pretrained LLM backbones to process continuous glucose monitoring time series through patch reprogramming.
Li et al. \cite{li2025llm} introduced Gluco-LLM, a personalized glucose-forecasting framework based on Time-LLM, in which a frozen pretrained LLM backbone is coupled with trainable time series reprogramming and output-projection modules to integrate CGM and patient-specific contextual information.
Gao et al. \cite{gao2026llm} proposed GlyLLM, an LLM-powered framework for personalized glycemic assessment in Type 2 Diabetes (T2D) that integrates continuous glucose monitoring data, wearable sensor signals, and patient-specific metadata to explore if incorporating personalized textual metadata together with wearable sensor information improves predictive performance over conventional ML baselines.
Mahmoudi et al. \cite{mahmoudi2026diabllm} proposed DiabLLM, an LLM-based framework for blood glucose prediction in T1D that adapts recent language-model-inspired forecasting architectures, namely Time-LLM and Chronos, to CGM data. Unlike prompt-based approaches, DiabLLM exploits pretrained LLM backbones for numerical time series forecasting.



Differently, Healey et al. \cite{healey2025case} investigated the use of GPT-4 for the analysis and summarization of continuous glucose monitoring (CGM) data rather than for glucose prediction, investigating if  LLMs can effectively transform raw physiological time series into clinically meaningful narrative descriptions. Healey and Kohane \cite{healey2024llm} proposed a benchmark for conversational analysis of CGM data that evaluates GPT-4-based text, code generation, and agentic frameworks on a range of objective glucose-related queries. He et al. \cite{he2026personalized} developed a personalized diabetes treatment-support system by fine-tuning compact general purpose LLMs with LoRA \cite{hu2021lora} on electronic health records. Ding et al. \cite{ding2024large} investigated how heterogeneous clinical information can be represented and integrated by language-model-based predictors, combining clinical notes with both numerical and textualized laboratory values for new-onset T2D prediction.

Taken together, these studies demonstrate the growing interest in LLM-based approaches to physiological time series analysis and glucose prediction, while adopting different strategies for representing and integrating physiological and contextual information. However, to the best of our knowledge, no prior work has investigated how the representation and type of physiological information affect prompt-based glucose prediction, particularly in the context of postprandial glucose forecasting in T1DM.
\section{Method}
\label{sec:method}

This work investigates postprandial glycemic-event prediction through the comparison of three distinct and complementary approaches. The first approach comprises conventional supervised ML models trained specifically for glucose prediction from patient-specific physiological observations. The second formulates glucose prediction as a prompt-based task performed by an unchanged general purpose LLM. The third approach is represented by Gluco-LLM \cite{li2025llm}.

Unlike supervised ML models, which learn a prediction function directly from labeled physiological data, and unlike Gluco-LLM, which adapts a pretrained LLM through dedicated trainable components, the chosen prompt-based framework intentionally leaves the underlying LLM unchanged throughout the entire experimental protocol. Consequently, any observed variation in predictive performance can be attributed to differences in the information provided to the model rather than to parameter optimization or architectural modifications. This design enables a principled investigation of prompt-based processing independently of task-specific model adaptation.

\begin{figure*}[t]
    \centering
    \includegraphics[width=\textwidth]{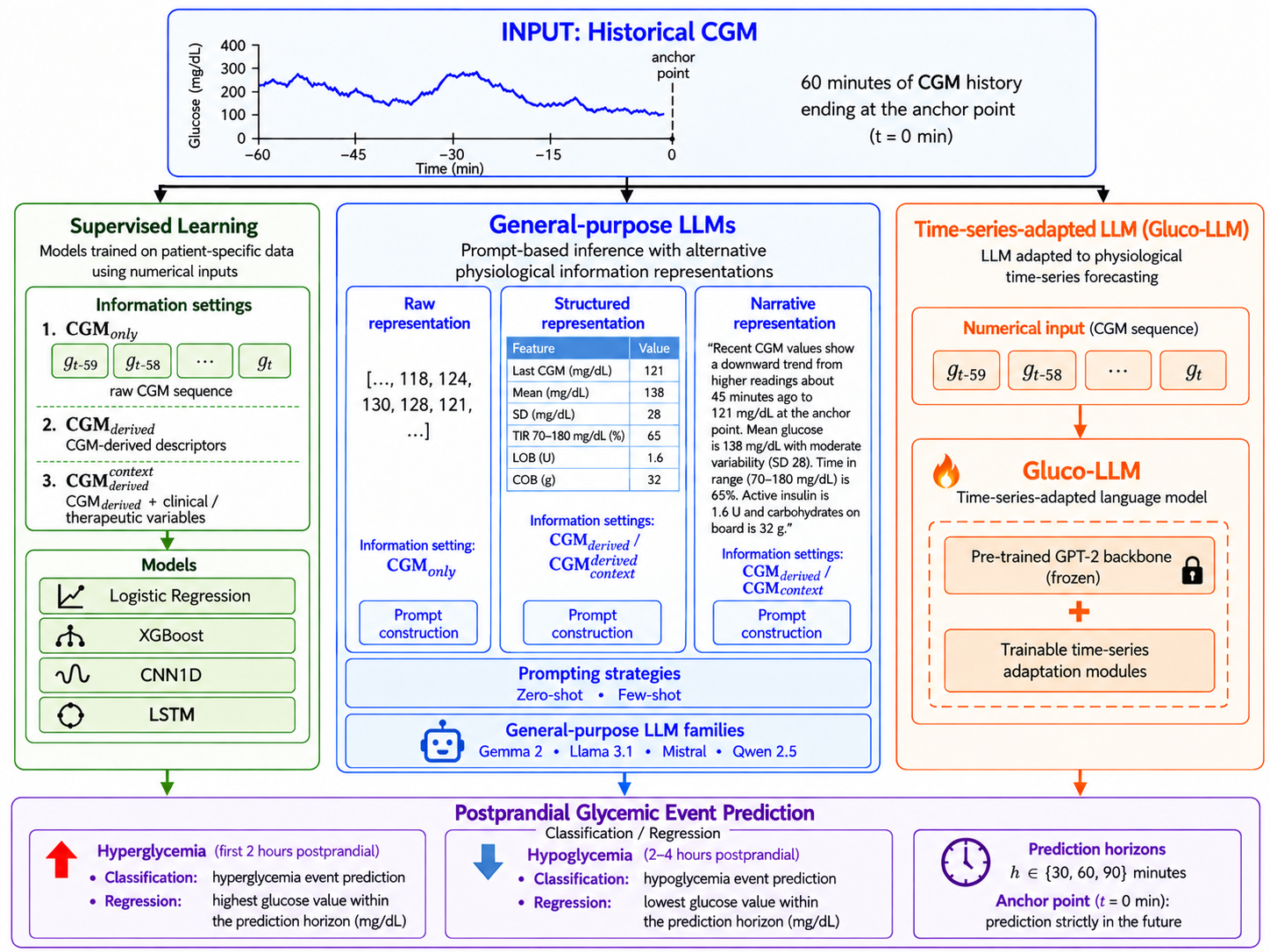}
    \caption{The framework compares three complementary approaches for postprandial glycemic-event prediction from continuous glucose monitoring (CGM) data: (i) conventional supervised ML models, (ii) prompt-based inference with a general purpose LLM using alternative physiological information representations, and (iii) Gluco-LLM \cite{li2025llm}, a recently proposed LLM-based architecture explicitly adapted to glucose time series forecasting. The chosen prompt-based framework provides the primary methodological basis for investigating the three research questions addressed in this work.}
    \label{fig:framework}
\end{figure*}

Figure~\ref{fig:framework} summarizes the overall experimental framework.
Starting from the same postprandial prediction setting, the three considered approaches are evaluated on aligned prediction tasks and horizons. Standard supervised models directly process numerical physiological inputs, whereas Gluco-LLM operates on numerical time series data through its dedicated time series adaptation pipeline. In contrast, the prompt-based configuration used in our experiments converts the available physiological information into a textual representation, which is then incorporated into a prompt and submitted to a general purpose LLM.

Within the prompt-based framework, two complementary aspects of the input are investigated. The first concerns physiological information representation, namely how the available patient observations are expressed before being presented to the language model. To investigate how the form in which physiological information is presented affects LLM-based prediction, we define three alternative representation strategies: \emph{Raw}, \emph{Structured}, and \emph{Narrative}. \emph{Raw} preserves the original CGM sequence through direct textual serialization, whereas \emph{Structured} and \emph{Narrative} transform derived physiological descriptors into, respectively, a key--value representation and a natural-language description.
The second aspect concerns the physiological information made available to the model. In particular, for \emph{Structured} and \emph{Narrative} representations, two information settings are evaluated: i)  $CGM_{derived}$ contains only descriptors derived exclusively from the historical CGM sequence, whereas ii) $CGM_{derived}^{context}$ augments these descriptors with additional clinical and therapeutic variables related to insulin administration, carbohydrate intake, basal insulin, meal timing, and physical activity. All these variables are summarized in Table \ref{tab:descriptors}.

Notice that the effect of linguistic representation can be examined by comparing  \emph{Structured} and \emph{Narrative} under the same information setting, whereas the contribution of additional physiological information can be assessed by comparing $CGM_{derived}$ and $CGM_{derived}^{context}$ within the same representation. Comparisons involving \emph{Raw} are complementary, since \emph{Raw} preserves the original CGM sequence whereas \emph{Structured} and \emph{Narrative} operate on descriptors derived from that sequence; such comparisons therefore jointly reflect differences in representation and information transformation.
Under this perspective, the prompt acts as the interface through which different representations of the available physiological information are communicated to the language model, while the underlying model parameters remain unchanged.

\subsection{Problem formulation}
\label{subsec:problem}

To address the research objectives of this work, we adopt a meal-centered formulation. Each meal defines a postprandial episode, and prediction samples are generated within the corresponding postprandial period. Following the protocol proposed by Cui et al.~\cite{cui2023jointly}, each meal defines a four-hour monitoring interval. The first two hours are considered for hyperglycemia prediction, whereas the following two hours are considered for hypoglycemia prediction. This temporal separation is consistent with the different temporal profiles of postprandial glycemic excursions: hyperglycemic events are more likely to occur during the early postprandial phase, while hypoglycemic events typically emerge later as the effects of insulin action become more prominent. This formulation reflects the objective of anticipating adverse glycemic events following meal intake, where glucose dynamics are strongly influenced by the interaction between carbohydrate absorption and insulin action.
As above discussed, the information made available to the prediction model may correspond either to the historical CGM sequence itself, to a set of descriptors derived from that sequence, or to the same descriptors augmented with additional contextual information.

We denote by $\mathcal{I}(s,t)$ the information available to the model for sample $t$ under information setting $s$. Specifically,

$
\mathcal{I}(s,t) =
\begin{cases}
X_t,
& s=CGM_{only},\\[4pt]
\phi(X_t),
& s=CGM_{derived},\\[4pt]
[\phi(X_t), C_t],
& s=CGM_{derived}^{context},
\end{cases}
$

where $X_t=\{g_{t-T+1},g_{t-T+2},\ldots,g_t\}$ denotes the historical CGM sequence available at prediction time $t$, $T$ is the observation-window length, and $g_i$ is the glucose concentration measured at time $i$.
The function $\phi(\cdot)$ maps the historical CGM window $X_t$ to a set of statistical and temporal descriptors computed exclusively from the CGM observations available up to time $t$. In contrast, $C_t$ denotes the collection of contextual variables available at prediction time $t$ that cannot be derived directly from $X_t$. Accordingly, the $CGM_{derived}^{context}$ setting combines the CGM-derived descriptors $\phi(X_t)$ with the additional contextual information $C_t$. The variables included in $\phi(X_t)$ and $C_t$ are summarized in Table~\ref{tab:information_settings}, while the corresponding preprocessing procedures are detailed in Section~\ref{subsec:dataset_preprocessing}.

The three feature settings were designed to capture complementary ways of making patient information available to the prediction model. The $CGM_{only}$ setting preserves the original temporal structure of the CGM signal and requires the model to infer relevant temporal patterns directly from the glucose trajectory.
By contrast, as already above discussed and summarized in Table \ref{tab:descriptors}, the $CGM_{derived}$ setting instead adopts only set of CGM-derived characteristics describing the patient's current glycemic state, recent temporal evolution, variability, and exposure to clinically relevant glucose ranges. Finally, the $CGM_{derived}^{context}$ setting complements these CGM-derived descriptors with contextual factors that may affect postprandial glucose dynamics but are not directly observable from CGM alone, including information related to insulin administration, carbohydrate intake, meal timing, and physical activity.

This design allows us to investigate whether LLM-based prediction benefits from direct access to the original CGM trajectory, from an explicit representation of characteristics derived from that trajectory, or from the availability of complementary contextual information beyond CGM 

\begin{table*}[t]
\centering
\caption{Information available at prediction time $t$ under the considered information settings. The $CGM_{derived}$ setting contains descriptors $\phi(X_t)$ derived from the historical CGM window $X_t$, whereas the $CGM_{derived}^{context}$ setting additionally includes contextual variables $C_t$ available at prediction time $t$.}
\label{tab:information_settings}
\scalebox{0.7}{
\begin{tabular}{llll}
\toprule
\textbf{Setting} &
\textbf{Attribute} &
\textbf{Notation} &
\textbf{Description} \\
\midrule

$CGM_{only}$ &
Historical CGM sequence &
$X_t=\{g_{t-T+1},\ldots,g_t\}$ &
CGM measurements collected during the historical observation window ending at $t$.
\\
\midrule

$CGM_{derived}$ &
Current glucose &
$g_t$ &
Most recent CGM measurement available at prediction time $t$.
\\

&
Mean glucose &
$\mu(X_t)$ &
Mean glucose concentration over the historical CGM window.
\\

&
Standard deviation &
$\sigma(X_t)$ &
Variability of glucose measurements within the historical window.
\\

&
Minimum glucose &
$\min(X_t)$ &
Minimum glucose concentration observed within the historical window.
\\

&
Maximum glucose &
$\max(X_t)$ &
Maximum glucose concentration observed within the historical window.
\\

&
Recent minimum &
$\min(X_t^{\mathrm{recent}})$ &
Minimum glucose concentration over the predefined recent portion of the historical window.
\\

&
Coefficient of variation &
$\mathrm{CV}(X_t)$ &
Glucose variability normalized with respect to the mean glucose level.
\\

&
Glucose slope &
$\mathrm{slope}(X_t)$ &
Estimated temporal trend of glucose measurements within the historical window.
\\

&
Most recent glucose change &
$g_t-g_{t-1}$ &
Difference between the two most recent CGM measurements.
\\

&
Time above range &
$\mathrm{TAR}(X_t)$ &
Fraction of the historical window spent above the target glucose range.
\\

&
Time below range &
$\mathrm{TBR}(X_t)$ &
Fraction of the historical window spent below the target glucose range.
\\

&
Time in range &
$\mathrm{TIR}(X_t)$ &
Fraction of the historical window spent within the target glucose range.
\\

&
Number of readings &
$|X_t|$ &
Number of valid CGM measurements available in the historical window.
\\

&
Window duration &
$\Delta(X_t)$ &
Temporal duration covered by the available CGM measurements.
\\
\midrule

$CGM_{derived}^{context}$ &
Insulin on board &
$\mathrm{IOB}_t$ &
Estimated amount of previously administered insulin still active at time $t$.
\\

&
Carbohydrates on board &
$\mathrm{COB}_t$ &
Estimated amount of previously ingested carbohydrates still being absorbed at time $t$.
\\

&
Basal insulin rate &
$\mathrm{Basal}_t$ &
Basal insulin delivery rate available at prediction time $t$.
\\

&
Number of boluses &
$N_{\mathrm{bolus},t}$ &
Number of bolus insulin administrations considered before prediction time $t$.
\\

&
Number of meals &
$N_{\mathrm{meal},t}$ &
Number of recorded meals considered before prediction time $t$.
\\

&
Time since last meal &
$\Delta t_{\mathrm{meal},t}$ &
Elapsed time between the most recent recorded meal and prediction time $t$.
\\

&
Physical-activity duration &
$A_t$ &
Duration of recent physical activity considered at prediction time $t$.
\\
\bottomrule
\end{tabular}}
\label{tab:descriptors}
\end{table*}

\subsubsection{Type of predictions}
The objectives of this study are addressed in the context of a postprandial glycemic event prediction task, formulated as a binary decision concerning the occurrence of hyperglycemia or hypoglycemia within a selected prediction horizon. Following Cui et al.~\cite{cui2023jointly}, two alternative formulations are investigated. In the direct-classification formulation, the available physiological information is mapped directly to a binary event prediction. In the regression-based formulation, the model first estimates a continuous future glucose outcome, from which the binary event prediction is subsequently obtained by applying the corresponding clinical threshold on the maximum
for hyperglycemia or the minimum for hypoglycemia. Depending on the considered approach, the relevant future glucose extreme, i.e. the maximum for hyperglycemia or the minimum for hypoglycemia, is either predicted directly or extracted from a predicted future glucose trajectory. 

We highlight that, as in Cui et al.~\cite{cui2023jointly}, this threshold-based event decision obtained from the predicted glucose extreme is not strictly equivalent to the direct-classification event definition, which requires at least two consecutive CGM readings beyond the corresponding threshold. The regression-based formulation should therefore be interpreted as an approximate event-detection formulation derived from the continuous prediction target rather than as an exact reformulation of the direct binary-classification task. However, this formulation enables performance to be assessed both at the continuous glucose level and, after thresholding, at the event-classification level. 

Given the physiological information $\mathcal{I}(s,t)$ available at prediction time $t$, the objective is to estimate whether an adverse glycemic event (namely, hyperglycemia and hypoglicemia) will occur within the selected prediction horizon $h$ for the classification problem, and the future glucose extreme occurring within the selected prediction horizon $h$ for the regression problem. For the classification problem, hyperglycemia and hypoglycemia classification target (i.e., the labels) are generated according to the postprandial prediction protocol proposed by Cui et al.~\cite{cui2023jointly}, assigning a positive label when at least two consecutive CGM readings within the prediction horizon are above 180 mg/dL or below 70 mg/dL, respectively, and a negative label otherwise.
Instead, for the regression problem, the regression target is defined as the maximum or the minimum CGM reading within the prediction horizon in the hyperglycemia and hypoglycemia regression problem, respectively.

To evaluate whether the continuous prediction also correctly identifies the corresponding clinical event, binary predictions are subsequently derived by applying the same glucose thresholds adopted in the direct classification setting, i.e. hyperglicemia if the predicted value is above 180 mg/dL and hypoglicemia if the predicted value is below 70 mg/dL.

This formulation therefore permits the models to be evaluated both in terms of similarity to the predicted glucose extreme and in terms of their ability to identify relevant glycemic events after thresholding.

For conventional supervised ML models, the information $\mathcal{I}(s,t)$ available for a patient sample is provided directly to the classifier. 
Differently, general purpose LLMs interact with the prediction task through a textual interface. Consequently, the available physiological information must be encoded into a compatible representation before being provided to the model. As above introduced, we investigate three alternative prompt representation modalities, referred to as \emph{Raw}, \emph{Structured}, and \emph{Narrative}. The \emph{Raw} representation preserves the historical CGM sequence through direct textual serialization, whereas the \emph{Structured} and \emph{Narrative} represent CGM-derived descriptors and, when available, additional contextual information through, respectively, explicit key-value pairs and natural language descriptions. Each prompt is generated deterministically from the physiological information available at the corresponding prediction time, using predefined templates and representation rules.
Furthermore, two inference conditions are considered, zero-shot inference ($\mathrm{ZS}$) and few-shot inference ($\mathrm{FS}$). Under $\mathrm{ZS}$, the prompt contains the task instructions and the target sample without labeled examples. Under $\mathrm{FS}$, the target query is preceded by $K$ labeled in-context demonstrations. 
Finally, the prompt reference answer is a binary label for classification and the corresponding future glucose extreme for regression. 

More formally, given an inference condition $c \in\{\mathrm{ZS},\mathrm{FS}\}$, a representation modality $r \in \{\mathrm{Raw},\mathrm{Structured},\mathrm{Narrative}\}$, an information setting $s \in \{CGM_{only}, CGM_{derived}, CGM_{derived}^{context}\}$, a prediction task $\tau\in \{\mathrm{Classification},\mathrm{Regression}\}$, a prediction horizon $h$, the prompt-generation function $P_t(r,s,\tau,h,c)$ constructs the textual input submitted to the LLM for the sample observed at time $t$. Details about the prompt function will be provided in the following of this section.
The resulting prediction is expressed as $\hat{y}_{t,h}=LLM\left(P_t(r,s,\tau,h,c)\right),$ where ($\mathrm{LLM}$) denotes the investigated general purpose LLM.

\subsubsection{Prompt construction function $P_t$}
\label{subsubsec:prompt_construction}

For a patient sample observed at time $t$, the prompt $P_t(r,s,\tau,h,c)$ submitted to the LLM is defined as $ P_t(r,s,\tau,h,c)= S \oplus \mathcal{E}(r,s,\tau,h,c) \oplus \mathcal{R}_r\left(\mathcal{I}(s,t)\right) \oplus Q(\tau,h),$ where $\oplus$ denotes textual concatenation, $\mathcal{I}(s,t)$ denotes the information available under information setting $s$ as discussed in Section \ref{subsec:problem}, $\mathcal{R}_r(\cdot)$ denotes its textual encoding according to the representation modality $r$, $Q(\tau,h)$ specifies the prediction task $\tau$ and the prediction horizon $h$, and $\mathcal{E}(r,s,\tau,h,c)$ accounts for the possible in-context examples included in the LLM prompt under the zero-shot/few-shot inference condition $c \in \{ZS, FS\}$. The system instruction $S$ is kept fixed across all experimental configurations:

\begin{quote}
\small
\texttt{You are an AI assistant in a clinic with the task of aiding doctors}\\
\texttt{by predicting glycemic events in Type 1 Diabetes patients based}\\
\texttt{on continuous glucose monitor data.}
\end{quote}

$\mathcal{R}_r$, $\mathcal{E}$, and $Q$ are defined in the following.

\paragraph{Raw Modality prompt component ($\mathcal{R}_\mathrm{Raw}$)}
With this representation modality, the LLM receives the original historical glucose observations.
For the Raw representation, $\mathcal{R}_\mathrm{Raw}$ is obtained by directly serializing the historical CGM sequence, without derived descriptors:

$
\mathcal{R}_\mathrm{Raw}(\mathrm{CGM})=\mathrm{serialize}\left(g_{t-T+1},\ldots,g_t\right).$

The $\mathrm{serialize}$ reports in a textual representation the CGM measurements in chronological order. The most recent measurement $g_t$, corresponding to the current glucose level, is additionally reported explicitly in the textual template. No feature extraction or clinical interpretation is performed. The resulting textual block returned by the $\mathrm{serialize}$ function follows the deterministic template

\begin{quote}\small
\texttt{CGM readings (last 60 minutes, every 5 minutes):}\\
\texttt{[$g_{t-T+1}$, ..., $g_t$] mg/dL}\\
\texttt{Current glucose: $g_t$ mg/dL}
\end{quote}

\paragraph{Structured modality prompt component ($\mathcal{R}_{\mathrm{Structured}}$)}

The Structured representation expresses the available patient information through explicit descriptors organized as key--value pairs. Unlike the \emph{Raw} representation, the language model does not need to infer basic descriptive properties directly from the glucose sequence, since relevant statistical quantities and, when available, additional contextual variables are explicitly provided. The key--value organization also makes the semantic role of each variable explicit, allowing the model to operate directly on higher-level descriptors of the patient's recent state.

For the \emph{Structured} representation, the textual encoding is obtained by applying the representation function $\mathcal{R}_{\mathrm{Structured}}$ to $\mathcal{I}(s,t)$:

$ \mathcal{R}_{\mathrm{Structured}}\!\left(\mathcal{I}(s,t)\right)
=\bigoplus_{j=1}^{m_s} \mathrm{format}(k_j,v_j), $

where $(k_j,v_j)$ denotes the $j$-th key--value pair associated with the selected information setting, $m_s$ is the corresponding number of variables, and $\oplus$ denotes textual concatenation.

The resulting textual representation given by the $\mathrm{format}(\cdot)$ function has the general form 
\begin{quote}
    attribute name: value
\end{quote}
Therefore, the final $\mathcal{R}_{\mathrm{Structured}}$ returns a textual representation of the form

\begin{quote}\small
\texttt{Patient: Type 1 Diabetes}\\
\texttt{Monitoring window: 60 minutes}\\
\texttt{Current glucose: [value] mg/dL}\\
\texttt{Mean glucose: [value] mg/dL}\\
\texttt{...}
\end{quote}

\paragraph{Narrative modality prompt component ($\mathcal{R}_{\mathrm{Narrative}}$)}
The Narrative representation expresses the information available under a given information setting through natural language sentences. For a fixed information setting $s \in \{CGM_{derived}, CGM_{derived}^{context}\}$, it is generated from the same underlying information $\mathcal{I}(s,t)$ used by the \emph{Structured} representation, differing only in the way this information is textually presented to the language model.

Formally, the \emph{Narrative} representation is obtained by applying a  verbalization function to the available information: 

$\mathcal{R}_{\mathrm{Narrative}}\!\left(\mathcal{I}(s,t)\right)= \mathrm{verbalize}\!\left(\mathcal{I}(s,t)\right)$. 

The function $\mathrm{verbalize}(\cdot)$ 
encodes the available variables as follows: 

\begin{quote}\small
\texttt{A patient with type 1 diabetes has been monitored during the last}\\
\texttt{60 minutes. The current glucose level is [value] mg/dL.}\\
\texttt{[Automatically generated physiological description]}
\end{quote}

Therefore, keys and their values are inserted into predefined natural language expressions and combined into coherent sentences describing the patient's recent glucose dynamics and, when available, the additional contextual information.

Thus, for a fixed information setting $s$, \emph{Structured} and \emph{Narrative} representations operate on the same underlying information
$\mathcal{I}(s,t)$ and differ only in their textual realization: the former uses explicit key--value pairs, whereas the latter expresses the same content through natural-language sentences.
\paragraph{Zero-shot and few-shot prompt component ($\mathcal{E}$)}
\label{subsubsec:in_context}

Prompt-based inference is investigated under both zero-shot (ZS) and few-shot (FS) conditions. In both cases, the parameters of the language model remain unchanged, and task-specific conditioning is provided exclusively through the prompt.

Following the prompt formalization introduced above, the inference condition is denoted by $c\in\{\mathrm{ZS},\mathrm{FS}\}$ and determines whether labeled in-context examples are included through the term $\mathcal{E}(r,s,\tau,h,c)$.

In the ZS condition, no labeled examples are provided and therefore $\mathcal{E}(r,s,\tau,h,\mathrm{ZS})$ returns the empty textual sequence. Consequently, the prompt reduces to $P_t(r,s,\tau,h,\mathrm{ZS})=S\oplus\mathcal{R}_r\!\left(\mathcal{I}(s,t)\right)\oplus Q(\tau,h)$.
The model is therefore required to perform the prediction without observing any labeled examples of the considered task.

In the FS condition, $\mathcal{E}(r,s,\tau,h,\mathrm{FS})$ contains $K$ labeled in-context examples selected according to the procedure described in Section~\ref{sec:experimental_assessment}. Each example is constructed using the same representation strategy $r$, information setting $s$, prediction task $\tau$, and prediction horizon $h$ adopted for the target sample, and includes the corresponding input representation together with its expected output.

Formally, $\mathcal{E}(r,s,\tau,h,\mathrm{FS}) = \bigoplus_{i=1}^{K} E_i(r,s,\tau,h),$ where $E_i(r,s,\tau,h)$ denotes the $i$-th labeled in-context example and $\oplus$ denotes textual concatenation. The corresponding few-shot prompt is $P_t(r,s,\tau,h,\mathrm{FS})=S\oplus\mathcal{E}(r,s,\tau,h,\mathrm{FS})\oplus\mathcal{R}_r\left(\mathcal{I}(s,t)\right)\oplus Q(\tau,h).$

For classification, the expected outputs associated with the in-context examples are \texttt{Yes} or \texttt{No}; whenever possible, positive and negative examples are balanced to avoid introducing an artificial preference toward either class. For regression, the expected output is the corresponding numerical glucose extreme expressed in mg/dL.
The resulting text from the $\mathcal{E}(r,s,\tau,h,\mathrm{FS})$ function when $r=\mathrm{Narrative}$ and $\tau=\mathrm{Classification}$ has the form:

\begin{quote}\small
\texttt{Example 1:}\\
\texttt{A patient with type 1 diabetes has been monitored during the last}\\
\texttt{60 minutes. The current glucose level is [value] mg/dL.}\\
\texttt{[Automatically generated physiological description]}\\
\texttt{Will this patient experience [hyperglycemia/hypoglycemia] in the}\\
\texttt{next [horizon] minutes?}\\
\texttt{Answer: [Yes/No]}\\[4pt]
\texttt{Example 2:}\\
\texttt{A patient with type 1 diabetes has been monitored during the last}\\
\texttt{60 minutes. The current glucose level is [value] mg/dL.}\\
\texttt{[Automatically generated physiological description]}\\
\texttt{Will this patient experience [hyperglycemia/hypoglycemia] in the}\\
\texttt{next [horizon] minutes?}\\
\texttt{Answer: [Yes/No]}\\[4pt]
\dots\\
\texttt{Example K:}\\
\texttt{A patient with type 1 diabetes has been monitored during the last}\\
\texttt{60 minutes. The current glucose level is [value] mg/dL.}\\
\texttt{[Automatically generated physiological description]}\\
\texttt{Will this patient experience [hyperglycemia/hypoglycemia] in the}\\
\texttt{next [horizon] minutes?}\\
\texttt{Answer: [Yes/No]}\\[4pt]
\texttt{Now consider the following case:}\\
\end{quote}
The same ZS and FS protocols are adopted for the regression task, with the task instruction and expected response adapted to the continuous prediction target. Specifically, whereas classification demonstrations associate each input with a binary \texttt{Yes}/\texttt{No} label, regression demonstrations associate it with the corresponding future glucose extreme expressed in mg/dL. In every configuration, the in-context demonstrations are constructed using the same physiological representation $(r)$ and information setting $(s)$ as the target query. Accordingly, demonstrations based on the \emph{Structured} and \emph{Narrative} representations are generated under either the $CGM_{derived}$ or $CGM_{derived}^{context}$ information setting, consistently with the evaluated query. The number (K) of in-context demonstrations and their selection procedure are detailed in the experimental setup, as they constitute experimental design choices rather than intrinsic components of the proposed framework.
\paragraph{Task-specific prompt component ($Q$)}
\label{subsubsec:Task-specific_component}

For the classification formulation, $Q(\tau,h)$ instructs the language model to determine whether a hyperglycemic or hypoglycemic event will occur within the specified prediction horizon and to answer exclusively with \texttt{Yes} or \texttt{No}. Restricting the expected output format enables deterministic mapping of the generated response to the corresponding binary prediction.

For the regression formulation, $Q(\tau,h)$ asks the model to estimate the relevant future glucose extreme within the selected prediction horizon.
Specifically, the requested output corresponds to the maximum glucose value for hyperglycemia prediction and to the minimum glucose value for hypoglycemia prediction. The model is instructed to return a single integer expressed in mg/dL, which is subsequently parsed for automatic evaluation.
For the classification formulation, the task-specific instruction is defined as

\begin{quote}\small
\texttt{Will this patient experience hyperglycemia in the next $h$ minutes?}\\
\texttt{Answer Yes or No.}
\end{quote}

for hyperglycemia prediction, and

\begin{quote}\small
\texttt{Will this patient experience hypoglycemia in the next $h$ minutes?}\\
\texttt{Answer Yes or No.}
\end{quote}

for hypoglycemia prediction.

For the regression formulation, the task-specific instruction is defined as

\begin{quote}\small
\texttt{Predict the maximum glucose value (mg/dL) the patient will reach}\\
\texttt{in the next $h$ minutes. Answer with a single integer.}
\end{quote}

for hyperglycemia prediction, and

\begin{quote}\small
\texttt{Predict the minimum glucose value (mg/dL) the patient will reach}\\
\texttt{in the next $h$ minutes. Answer with a single integer.}
\end{quote}

for hypoglycemia prediction.

In Tab. \ref{tab:prompt_summary} the prompt construction is summarized.

\begin{table*}[!t]
\centering
\scriptsize
\caption{Representative prompt structures obtained by combining the common system instruction $S$, the in-context component $\mathcal{E}$, the task-specific component $Q$, and the representation component $\mathcal{R}_r$. The table shows hyperglycemia prediction at a generic horizon $h$. Raw prompts use the $CGM_{only}$ information setting,
whereas \emph{Structured} and \emph{Narrative} prompts are illustrated under
$CGM_{derived}$. Under the $CGM_{derived}^{context}$ setting, the same templates
additionally include the available contextual variables. In few-shot prompts,
only the first of the $K$ demonstrations is shown explicitly.}
\label{tab:prompt_summary}

\setlength{\tabcolsep}{3pt}
\renewcommand{\arraystretch}{1.08}

\newcommand{\SystemInstruction}{%
\parbox[c]{0.68\textwidth}{%
\centering
\texttt{You are an AI assistant in a clinic with the task of aiding doctors}\\
\texttt{by predicting glycemic events in Type 1 Diabetes patients based}\\
\texttt{on continuous glucose monitor data.}
}}

\newcommand{\RawBlock}{%
\texttt{CGM readings (last 60 minutes, every 5 minutes):}\par
\texttt{[$g_{t-T+1}$, \ldots, $g_t$] mg/dL}\par
\texttt{Current glucose: $g_t$ mg/dL}\par
}

\newcommand{\StructuredBlock}{%
\texttt{Patient: Type 1 Diabetes}\par
\texttt{Monitoring window: 60 minutes}\par
\texttt{Current glucose: [value] mg/dL}\par
\texttt{Mean glucose: [value] mg/dL}\par
\texttt{[Additional CGM-derived descriptors]}\par
}

\newcommand{\NarrativeBlock}{%
\texttt{A patient with type 1 diabetes has been monitored during}\par
\texttt{the last 60 minutes. The current glucose level is}\par
\texttt{[value] mg/dL. [Physiological description]}\par
}

\newcommand{\ClassificationQuestion}{%
\texttt{Will this patient experience hyperglycemia in the next}\par
\texttt{$h$ minutes? Answer Yes or No.}\par
}

\newcommand{\RegressionQuestion}{%
\texttt{Predict the maximum glucose value (mg/dL) the patient}\par
\texttt{will reach in the next $h$ minutes. Answer with a}\par
\texttt{single integer.}\par
}

\newcommand{\PromptCell}[1]{%
\begin{minipage}[t]{\linewidth}
\raggedright
#1
\end{minipage}
}

\newcommand{\ZSClassificationCell}[1]{%
\PromptCell{%
#1
\ClassificationQuestion
}}

\newcommand{\ZSRegressionCell}[1]{%
\PromptCell{%
#1
\RegressionQuestion
}}

\newcommand{\FSClassificationCell}[1]{%
\PromptCell{%
\texttt{Example 1:}\par
#1
\ClassificationQuestion
\texttt{Answer: [Yes/No]}\par
$\vdots$\par
\texttt{Example K: [same format]}\par
\texttt{Now consider the following case:}\par
#1
\ClassificationQuestion
}}

\newcommand{\FSRegressionCell}[1]{%
\PromptCell{%
\texttt{Example 1:}\par
#1
\RegressionQuestion
\texttt{Answer: [integer]}\par
$\vdots$\par
\texttt{Example K: [same format]}\par
\texttt{Now consider the following case:}\par
#1
\RegressionQuestion
}}

\scalebox{0.95}{%
\begin{tabularx}{\textwidth}{
    >{\centering\arraybackslash}m{0.8cm}
    >{\centering\arraybackslash}m{1.7cm}
    >{\raggedright\arraybackslash}X
    >{\raggedright\arraybackslash}X
    >{\raggedright\arraybackslash}X}
\hline

\multicolumn{2}{c}{\textbf{Common component $S$}}
&
\multicolumn{3}{c}{\SystemInstruction}
\\
\hline

\multirow[c]{2}{*}{$\mathcal{E}$}
&
\multirow[c]{2}{*}{$Q$}
&
\multicolumn{3}{c}{$\mathcal{R}_r$}
\\
\cline{3-5}

&
&
\multicolumn{1}{c}{\textbf{Raw}}
&
\multicolumn{1}{c}{\textbf{Structured}}
&
\multicolumn{1}{c}{\textbf{Narrative}}
\\
\hline

\multirow[c]{2}{*}{\textbf{ZS}}
&
Classification
&
\ZSClassificationCell{\RawBlock}
&
\ZSClassificationCell{\StructuredBlock}
&
\ZSClassificationCell{\NarrativeBlock}
\\
\cline{2-5}

&
Regression
&
\ZSRegressionCell{\RawBlock}
&
\ZSRegressionCell{\StructuredBlock}
&
\ZSRegressionCell{\NarrativeBlock}
\\
\hline

\multirow[c]{2}{*}{\textbf{FS}}
&
Classification
&
\FSClassificationCell{\RawBlock}
&
\FSClassificationCell{\StructuredBlock}
&
\FSClassificationCell{\NarrativeBlock}
\\
\cline{2-5}

&
Regression
&
\FSRegressionCell{\RawBlock}
&
\FSRegressionCell{\StructuredBlock}
&
\FSRegressionCell{\NarrativeBlock}
\\
\hline

\end{tabularx}%
}
\end{table*}

\subsubsection{Output parsing}
\label{subsubsec:output_parsing}

LLM responses are automatically converted into the output format required for quantitative evaluation. The parsing procedure depends on the considered prediction formulation and is applied consistently across all models and experimental conditions.

For direct classification, the model is instructed to return exclusively \texttt{Yes} or \texttt{No}. The resulting response is mapped to the corresponding positive or negative class, respectively, enabling direct comparison with the predictions produced by supervised classifiers.

For regression, the model is instructed to return a numerical estimate of the future glucose extreme, expressed in mg/dL. Since LLMs may occasionally include additional text in their response, the generated output is automatically parsed to recover the predicted glucose value. Specifically, the first valid numerical value occurring in the generated response is extracted and interpreted as the predicted glucose value. If the model returns a numerical interval, its midpoint is used as the prediction. Responses from which no valid numerical value can be recovered are treated as invalid predictions.  If no valid glucose value can be identified, the response is considered inconclusive and excluded from the evaluation.

The parsed regression output is evaluated directly as a continuous glucose prediction and is additionally converted into a binary event prediction by applying the corresponding clinical threshold. This enables the regression formulation to be assessed both in terms of numerical prediction accuracy and glycemic-event detection.

\subsection{Time series-adapted LLM: Gluco-LLM}
\label{subsec:glucollm}

In addition to prompt-based inference with general purpose LLMs, this study considers Gluco-LLM \cite{li2025llm}, an approach specifically designed to adapt pretrained language models to glucose time series forecasting. Gluco-LLM builds upon the Time-LLM framework~\cite{jin2024time} and represents a fundamentally different use of language models from the prompt-based approach considered in the previous sections. Rather than expressing physiological observations as textual inputs and querying an unchanged general purpose LLM, Gluco-LLM introduces trainable components that map numerical time series information into the representation space of a pretrained language-model backbone.

The historical CGM sequence is first divided into patches and mapped into an embedding space through a trainable patch-embedding module. A trainable mechanism subsequently aligns these time series representations with the embedding space of a pretrained GPT-2 backbone. The parameters of the language-model backbone remain frozen, whereas the adaptation components surrounding it are optimized for the target forecasting task. Finally, a trainable output projection maps the resulting hidden representation to the required numerical glucose prediction.

Gluco-LLM therefore provides an alternative approach between conventional task-specific forecasting models and prompt-based general purpose LLMs: it exploits a pretrained language-model backbone while learning dedicated interfaces for physiological time series prediction. 

In the present study, Gluco-LLM is adapted to the postprandial prediction formulation introduced in Section~\ref{subsec:problem}. Rather than forecasting glucose over the complete CGM trajectory, the model estimates the future glucose extreme within the considered prediction horizon, i.e., the maximum value for hyperglycemia and the minimum value for hypoglycemia. The corresponding binary event prediction is subsequently obtained by applying the same clinical thresholds adopted throughout this study.

Its specific training configuration and the protocol adopted for comparison with the supervised regression baseline are described in the experimental setup.

\section{Experimental assessment}
\label{sec:experimental_assessment}

The experimental assessment is designed to evaluate the three research questions introduced in Section~\ref{sec:introduction} under a common and controlled evaluation framework. All prediction approaches are evaluated on the same postprandial glucose-prediction setting, while the specific model configurations and comparison protocols are adapted to the objective of each research question.

The following sections first describe the dataset, preprocessing pipeline, input configurations, and evaluation criteria shared across the experiments. The experimental protocols adopted to address each research question are then presented separately.

\subsection{Dataset and preprocessing}
\label{subsec:dataset_preprocessing}

Experiments are conducted on the OhioT1DM dataset \cite{marling2020ohiot1dm}, a publicly available benchmark for glucose prediction in individuals with T1DM. The dataset contains physiological data from 12 patients monitored over several weeks. Following the postprandial protocol of Cui et al.~\cite{cui2023jointly}, subject 567 is excluded because no meal information is available in the official test partition. The final experimental cohort therefore consists of 11 patients. CGM measurements are recorded at five-minute intervals and are accompanied by additional patient information, including insulin administration, meal-related carbohydrate intake, fingerstick glucose measurements, and physical activity. 
Throughout this work, the historical observation window spans 60 minutes, resulting in $T=12$ measurements. \begin{table}[t]
\centering
\small
\caption{Number of prediction instances from the OhioT1DM dataset, reported for the training and test splits for each task and prediction horizon.}
\label{tab:dataset}

\setlength{\tabcolsep}{8pt}
\scalebox{0.9}{
\begin{tabular}{llrr}
\hline
\textbf{Task} &
\textbf{Horizon} &
\textbf{Training samples} &
\textbf{Test samples} \\
\hline

\multirow{3}{*}{Hyperglycemia}
& 30 min & 16\,384 & 3\,211 \\
& 60 min & 16\,194 & 3\,198 \\
& 90 min & 16\,011 & 3\,154 \\

\hline

\multirow{3}{*}{Hypoglycemia}
& 30 min & 25\,106 & 5\,560 \\
& 60 min & 16\,984 & 3\,739 \\
& 90 min & 8\,620 & 1\,896 \\

\hline
\end{tabular}}
\end{table}
The data used in this study are summarized in Table~\ref{tab:dataset}.
The prediction windows exhibit a marked class imbalance, reflecting the uneven occurrence of postprandial glycemic events. This imbalance is substantially more pronounced for hypoglycemia, whose positive instances remain rare across prediction horizons, whereas hyperglycemic events occur more frequently. Moreover, class prevalence varies with the prediction horizon, as longer horizons increase the probability of observing a threshold-crossing event. This characteristic is taken into account both during model training and in the selection of the evaluation metrics.

Data are processed independently for each patient. CGM measurements are ordered chronologically and represented on their nominal five-minute sampling grid following the protocol used in \cite{cui2023jointly}. Short missing intervals of up to 15 minutes are handled through forward filling, thereby relying exclusively on previously observed glucose values. Longer gaps are left missing, and any observation window containing such gaps is discarded. 


For input variables requiring normalization, scaling parameters are estimated independently from the training data of each patient and subsequently applied unchanged to validation and test observations. No statistics computed from the validation or test sets are therefore used during preprocessing of the training data \cite{apicella2025don}.

Postprandial samples are subsequently extracted according to the meal-centered protocol described in Section~\ref{subsec:problem}. Each prediction instance contains the previous 60 minutes of CGM history, corresponding to 12 measurements. Cui et al.~\cite{cui2023jointly} originally considered a 30 minute prediction horizon. In the present study, we retain their meal-centered extraction principle while extending the prediction horizon to $h\in\{30,60,90\}$ minutes. For each horizon, prediction instances are generated only when the complete future interval required to define the target is available within the corresponding task-specific postprandial region. Consequently, the latest admissible prediction time depends on $h$, resulting in a decreasing number of valid prediction instances as the prediction horizon increases.


\subsection{Compared models}
\label{subsec:compared_models}

The experimental assessment involves models belonging to the three prediction approaches introduced in Section~\ref{sec:method}: conventional supervised learning, prompt-based inference with general-purpose LLMs, and time series adaptation of a pretrained language model. Since the considered research questions address different prediction formulations, the role of the supervised reference models differs across the experimental protocols.

\paragraph{Supervised classification baselines}

For direct glycemic-event classification experiments, four supervised models are considered: Logistic Regression, XGBoost, a one-dimensional Convolutional Neural Network (CNN1D), and a Long Short-Term Memory network (LSTM) \cite{bishop2024deep}. The selected models cover complementary inductive biases, including a linear classifier, a nonlinear tree-based ensemble, and two neural architectures capable of modeling temporal patterns in glucose observations.

To account for class imbalance during the training of the neural models, we adopt a weighted binary cross-entropy loss \cite{phan2020resolving}. 
The weight is estimated exclusively from the training partition for each patient, prediction task, and prediction horizon, thereby preventing information from the validation or test sets from influencing model fitting. This weighting increases the penalty assigned to misclassified positive instances and counteracts the tendency to favor the majority negative class, which is particularly relevant for hypoglycemia prediction.

All four models are trained separately for each patient and evaluated on the same patient-specific test samples used for the corresponding prompt-based LLM experiments. Depending on the considered input configuration, they receive either the original CGM sequence or the engineered descriptors introduced in Section~\ref{sec:method}.

\paragraph{Regression reference model}

For the regression-based experiments, the supervised reference is the LSTM model following the postprandial prediction setting proposed by Cui et al.~\cite{cui2023jointly}. Unlike the direct classification baselines, this model predicts a continuous glucose target, namely the maximum future glucose value for hyperglycemia and the minimum future glucose value for hypoglycemia. The corresponding binary event prediction is subsequently obtained by applying the clinical thresholds defined in Section~\ref{subsec:problem}.

The Cui et al. \cite{cui2023jointly} model is employed as the supervised reference model both for the comparison with prompt-based LLM regression and for the evaluation of Gluco-LLM. This choice provides a common reference across the two regression-based experiments and enables the different LLM paradigms to be compared against the same task-specific forecasting approach.

\paragraph{General purpose LLMs}

Prompt-based experiments are conducted using four open-weight instruction-tuned language models from different model families: \emph{Llama~3.1~8B}, \emph{Mistral~7B}, \emph{Qwen~2.5~7B}, and \emph{Gemma~2~9B}. Considering multiple model families reduces the risk of attributing model-specific behavior to the prompt-based paradigm as a whole.

All models are executed locally through Ollama\footnote{https://ollama.com/} and are used without task-specific parameter updates. Prediction is therefore performed exclusively through the ZS and FS inference procedures described in Section~\ref{subsubsec:in_context}. To ensure a consistent inference setting, the context window is limited to 4096 tokens for all considered LLMs.

\paragraph{Time series-adapted LLM}
Gluco-LLM represents the time series-adapted LLM paradigm. As described in Section~\ref{subsec:glucollm}, the model combines a frozen GPT-2 backbone with trainable adaptation components specifically designed to map physiological time series data into the language-model representation space. Gluco-LLM is evaluated exclusively within the regression formulation and compared directly with the Cui et al. LSTM under an aligned postprandial prediction protocol.

This subsection describes the model selection criteria and implementation choices adopted for the evaluated predictive approaches. We first detail the conventional supervised models and then present the implementation of the general purpose and time series-adapted LLMs, including their training or inference configurations where applicable.

\paragraph{Supervised classification models}

For the direct-classification experiment, model selection is performed independently of the test set. For each patient, the last 15\% of the chronological training partition is used as a validation set, preserving temporal order. Hyperparameters are selected through grid search \cite{dangeti2017statistics} by maximizing the Matthews Correlation Coefficient (MCC, \cite{chicco2020advantages}) on the validation data.

The grid search hyperparameter space is summarized in Table~\ref{tab:hyperparameters}  
To account for variability due to stochastic optimization and initialization, the neural and stochastic supervised models are evaluated over multiple random seeds, and results are reported as mean and standard deviation where applicable.

\begin{table}[t]
\centering
\caption{Hyperparameter search spaces considered for the supervised classification models.}
\label{tab:hyperparameters}
\small
\scalebox{0.9}{
\begin{tabular}{lll}
\hline
\textbf{Model} & \textbf{Hyperparameter} & \textbf{Search space} \\
\hline

\multirow{6}{*}{CNN1D}
& Number of filters & $\{16,32,64,128\}$ \\
& Kernel size       & $\{2,3,5,6\}$ \\
& Dropout           & $\{0.2,0.5\}$ \\
& Learning rate     & $\{0.001,0.005,0.01,0.05,0.1\}$ \\
& Batch size        & $\{64\}$ \\
& Maximum epochs    & $\{50\}$ \\
\hline

\multirow{6}{*}{LSTM}
& Hidden size       & $\{32,64,128\}$ \\
& Number of layers  & $\{2,4,5,7\}$ \\
& Dropout           & $\{0.2,0.5\}$ \\
& Learning rate     & $\{0.001,0.005,0.01,0.05,0.1\}$ \\
& Batch size        & $\{64\}$ \\
& Maximum epochs    & $\{50\}$ \\
\hline

\multirow{3}{*}{XGBoost}
& Number of estimators & $\{100,200\}$ \\
& Maximum depth        & $\{4,6\}$ \\
& Learning rate        & $\{0.005,0.01,0.05\}$ \\
\hline

Logistic Regression
& Maximum iterations & $\{500,1000\}$ \\
\hline

\end{tabular}}
\end{table}
\paragraph{Prompt-based LLMs}

Prompt-based LLMs require neither training nor hyperparameter tuning. All four models are executed locally through Ollama under a common inference setup. For all models, the generation temperature is set to $0$, the maximum number of generated tokens is limited to $20$, and the context-window size is fixed to $4096$ tokens. The same prompt-generation, output-parsing, and ZS/FS procedures are applied consistently across model families. In the FS condition, For each target prediction, $K=6$ examples (Whenever possible, three positive and three negative)  are selected exclusively from the corresponding patient-specific training partition, thereby preventing test information from entering the prompt.    

\paragraph{Regression-based models}

The regression experiments follow a different model selection protocol in order to remain aligned with the reference postprandial forecasting setting. In particular, the Cui et al. model \cite{cui2023jointly} and Gluco-LLM are trained according to the optimization protocols adopted in their respective original studies.


\subsection{Evaluation metrics}
\label{subsec:evaluation_metrics}

Given the class imbalance characterizing the prediction problems, the MCC is adopted as the primary metric for model comparison, as it accounts for all four entries of the confusion matrix and provides a balanced assessment even when the class distribution is highly uneven.  MCC ranges from $-1$ to $+1$. Positive values, particularly those approaching $+1$, indicate increasingly good predictive performance, whereas values close to or below zero indicate poor performance.

For the regression formulation, prediction performance is evaluated using Root Mean Squared Error (RMSE) \cite{bishop2024deep} between the predicted and actual future glucose extreme. For hyperglycemia, the target corresponds to the maximum glucose value within the prediction horizon, whereas for hypoglycemia it corresponds to the minimum. Lower RMSE therefore indicates more accurate prediction of the future glucose extreme.

Additionally, the predicted continuous value is converted into a binary glycemic-event prediction using the same thresholds adopted for direct classification. The resulting decisions are evaluated using the MCC. RMSE and MCC therefore capture two complementary aspects of performance: numerical accuracy in predicting the future glucose extreme and the ability to correctly identify clinically relevant events.

For stochastic supervised and time series-adapted models evaluated over multiple runs, results are reported as mean and standard deviation across runs, consistently with several previous glucose-prediction studies~\cite{martinsson2020blood,cui2023jointly,voegeli2025beyond}.

Prompt-based general-purpose LLMs are instead evaluated using a single inference run for each experimental configuration and are therefore reported without an across-run standard deviation.

\subsection{Experimental protocols}
\label{subsec:experimental_protocols}
In this section, the experimental protocols adopted to address each research question are described.

\subsubsection{RQ1: General purpose LLMs vs.\ supervised learning}
\label{subsec:rq1_protocol}

RQ1 investigates whether prompt-based inference with general purpose LLMs can provide predictive performance comparable to conventional supervised ML models for postprandial glycemic event prediction. The primary comparison is conducted under the direct binary classification formulation introduced in Section~\ref{subsec:problem}.
The four supervised classification baselines, namely Logistic Regression, XGBoost, CNN1D, and LSTM, are compared with the four general purpose LLMs described in Section~\ref{subsec:compared_models}. 
The supervised baselines are trained separately for each patient. Their reported performance is obtained over 10 random seeds and summarized through mean and standard deviation. Prompt-based LLMs do not undergo task-specific training and are reported as a single prediction run for each zero-shot or few-shot experimental configuration.

\subsubsection{RQ2: Influence of physiological information representation and availability}
\label{subsec:rq2_protocol}

Since RQ2 investigates how the representation and the types of information made available to the model influence prompt-based prediction with general purpose LLMs,  this analysis focuses exclusively on the four prompt-based LLMs and examines the sensitivity of their predictions to alternative ways of presenting the patient information.

The effect of representation is first investigated by comparing the \emph{Raw}, \emph{Structured}, and \emph{Narrative} strategies introduced in Section~\ref{sec:method}. 
A second analysis investigates the effect of providing different kind of information, i.e., performance under the $CGM_{derived}$ configuration is compared with the corresponding $CGM_{derived}^{context}$ configuration in \emph{Structured} and \emph{Narrative} representations. The former contains only information derived from the CGM observations, whereas the latter additionally incorporates the clinical and therapeutic context described in Section~\ref{sec:method}. Comparing these configurations therefore isolates the effect of enriching the model input with information beyond glucose measurements.

\subsubsection{RQ3: Prompt-based inference vs.\ time series-adapted LLMs}
\label{subsec:rq3_protocol}

RQ3 investigates whether prompt-based inference with general purpose LLMs can approach the predictive performance of an LLM architecture explicitly adapted to physiological time series forecasting. To establish a meaningful comparison between these two paradigms, the analysis is conducted under the regression formulation introduced in Section~\ref{subsec:problem}, using the model proposed by Cui et al.~\cite{cui2023jointly} as a common supervised reference. 

The first stage compares the four general purpose LLMs with the Cui et al. model. We highlight that the primary comparison for RQ3 concerns the relative performance of the two LLM paradigms, while the Cui et al. model provides a common task-specific reference. Both approaches are required to estimate the future glucose extreme, namely the maximum glucose value for hyperglycemia and the minimum glucose value for hypoglycemia, at the different investigated prediction horizons. To preserve input comparability, prompt-based LLMs are evaluated exclusively using the \emph{Raw} representation of the CGM sequence, without engineered descriptors or additional clinical information. Each LLM is evaluated under both XS and FS inference.

The Cui et al. model is trained separately for each patient on the complete patient-specific training partition following the training protocol adopted in the reference setting. Its results are reported as mean and standard deviation over 10 initialization seeds, whereas each prompt-based LLM configuration produces a single inference result.

The second stage evaluates Gluco-LLM against the same Cui et al. model as reference. For this comparison, the Gluco-LLM data-loading procedure is explicitly aligned with that of the Cui et al. model so that both models operate on the same postprandial samples, observation windows, regression targets, and chronological data partitions. 
Performance is evaluated both on the continuous glucose prediction, through RMSE, and on the corresponding threshold-derived event prediction, through MCC.
\section{Results}
\label{sec:results}
\begin{table*}[t]
\centering
\small
\caption{MCC obtained by the supervised models across prediction tasks, input configurations, and prediction horizons. Results for stochastic models are reported as mean (standard deviation) over 10 runs. Bold values indicate the best supervised result for each task, input configuration, and prediction horizon.}
\label{tab:results_supervised}
\setlength{\tabcolsep}{7pt}
\begin{tabular}{lllcccc}
\hline
\textbf{Task} &
\textbf{Input} &
\textbf{Horizon} &
\textbf{LogReg} &
\textbf{XGBoost} &
\textbf{CNN1D} &
\textbf{LSTM} \\
\hline

\multirow{9}{*}{Hyperglycemia}
& \multirow{3}{*}{$CGM_{only}$}
& 30 min & \textbf{0.558} & 0.511 (0.003) & 0.454 (0.009) & 0.505 (0.028) \\
& & 60 min & \textbf{0.459} & 0.382 (0.004) & 0.296 (0.038) & 0.391 (0.018) \\
& & 90 min & \textbf{0.389} & 0.301 (0.004) & 0.236 (0.036) & 0.308 (0.024) \\
\cline{2-7}

& \multirow{3}{*}{$CGM_{derived}$}
& 30 min & \textbf{0.570} & 0.540 (0.005) & 0.475 (0.013) & 0.411 (0.074) \\
& & 60 min & \textbf{0.456} & 0.406 (0.005) & 0.377 (0.013) & 0.365 (0.028) \\
& & 90 min & \textbf{0.366} & 0.318 (0.002) & 0.271 (0.047) & 0.253 (0.060) \\
\cline{2-7}

& \multirow{3}{*}{$CGM_{derived}^{context}$}
& 30 min & 0.560 & \textbf{0.565 (0.005)} & 0.431 (0.011) & 0.364 (0.020) \\
& & 60 min & \textbf{0.523} & 0.497 (0.004) & 0.309 (0.074) & 0.329 (0.076) \\
& & 90 min & 0.403 & \textbf{0.455 (0.004)} & 0.274 (0.088) & 0.313 (0.066) \\
\hline

\multirow{9}{*}{Hypoglycemia}
& \multirow{3}{*}{$CGM_{only}$}
& 30 min & 0.331 & \textbf{0.350 (0.006)} & 0.231 (0.022) & 0.230 (0.050) \\
& & 60 min & \textbf{0.328} & 0.305 (0.008) & 0.202 (0.023) & 0.233 (0.016) \\
& & 90 min & \textbf{0.312} & 0.290 (0.002) & 0.179 (0.015) & 0.262 (0.050) \\
\cline{2-7}

& \multirow{3}{*}{$CGM_{derived}$}
& 30 min & 0.375 & \textbf{0.383 (0.003)} & 0.270 (0.085) & 0.210 (0.031) \\
& & 60 min & \textbf{0.345} & 0.330 (0.001) & 0.316 (0.039) & 0.232 (0.033) \\
& & 90 min & 0.351 & \textbf{0.366 (0.002)} & 0.283 (0.056) & 0.229 (0.031) \\
\cline{2-7}

& \multirow{3}{*}{$CGM_{derived}^{context}$}
& 30 min & 0.296 & \textbf{0.311 (0.008)} & 0.225 (0.092) & 0.129 (0.023) \\
& & 60 min & 0.280 & \textbf{0.308 (0.008)} & 0.302 (0.028) & 0.186 (0.043) \\
& & 90 min & 0.291 & \textbf{0.321 (0.003)} & 0.257 (0.051) & 0.190 (0.022) \\
\hline
\end{tabular}
\end{table*}
\begin{table*}[t]
\centering
\small
\caption{MCC of the general-purpose LLMs under zero-shot inference. For \emph{Structured} and \emph{Narrative} representations, values are reported as
$CGM_{derived}$ / $CGM_{derived}^{context}$. \emph{Raw} corresponds
to the CGM configuration. Bold values indicate the best input configuration for each LLM, prediction task, and horizon. A dash denotes an undefined MCC due to a constant prediction output.}
\label{tab:results_zs}
\setlength{\tabcolsep}{6pt}

\begin{tabular}{lllcccc}
\hline
\textbf{Task} &
\textbf{Horizon} &
\textbf{Representation} &
\textbf{Gemma 2} &
\textbf{Llama 3.1} &
\textbf{Mistral} &
\textbf{Qwen 2.5} \\
\hline

\multirow{9}{*}{Hyperglycemia}
& \multirow{3}{*}{30 min}
& Raw
& 0.298 & 0.254 & 0.335 & 0.367 \\
&
& Structured
& 0.376 / \textbf{0.392}
& \textbf{0.357} / 0.346
& 0.270 / \textbf{0.408}
& 0.372 / \textbf{0.485} \\
&
& Narrative
& 0.204 / 0.202
& 0.227 / 0.200
& 0.238 / 0.214
& 0.314 / 0.247 \\
\cline{2-7}

& \multirow{3}{*}{60 min}
& Raw
& 0.299 & 0.244 & 0.310 & 0.305 \\
&
& Structured
& 0.357 / \textbf{0.369}
& \textbf{0.329} / 0.313
& 0.256 / \textbf{0.360}
& 0.398 / \textbf{0.407} \\
&
& Narrative
& 0.168 / 0.162
& 0.205 / 0.182
& 0.200 / 0.184
& 0.280 / 0.204 \\
\cline{2-7}

& \multirow{3}{*}{90 min}
& Raw
& 0.271 & 0.206 & \textbf{0.284} & 0.240 \\
&
& Structured
& 0.331 / \textbf{0.338}
& \textbf{0.290} / \textbf{0.290}
& 0.189 / 0.265
& 0.331 / \textbf{0.375} \\
&
& Narrative
& 0.139 / 0.126
& 0.172 / 0.140
& 0.151 / 0.146
& 0.245 / 0.194 \\
\hline

\multirow{9}{*}{Hypoglycemia}
& \multirow{3}{*}{30 min}
& Raw
& 0.341 & 0.088 & \textbf{0.274} & \textbf{0.421} \\
&
& Structured
& 0.458 / \textbf{0.507}
& -- / --
& -- / -0.002
& -0.002 / -0.004 \\
&
& Narrative
& 0.132 / 0.140
& 0.270 / \textbf{0.271}
& 0.198 / 0.137
& 0.377 / 0.314 \\
\cline{2-7}

& \multirow{3}{*}{60 min}
& Raw
& 0.373 & 0.092 & 0.270 & \textbf{0.429} \\
&
& Structured
& \textbf{0.456} / 0.403
& -- / --
& -- / -0.004
& -0.003 / -0.008 \\
&
& Narrative
& 0.183 / 0.195
& \textbf{0.336} / 0.318
& \textbf{0.283} / 0.200
& 0.345 / 0.282 \\
\cline{2-7}

& \multirow{3}{*}{90 min}
& Raw
& 0.386 & 0.073 & \textbf{0.337} & \textbf{0.452} \\
&
& Structured
& \textbf{0.436} / 0.406
& -- / --
& -- / -0.005
& -0.005 / -0.011 \\
&
& Narrative
& 0.240 / 0.247
& \textbf{0.369} / 0.357
& 0.315 / 0.264
& 0.376 / 0.322 \\
\hline
\end{tabular}
\end{table*}

\begin{table*}[t]
\centering
\small
\caption{MCC of the general-purpose LLMs under few-shot inference. For
\emph{Structured} and \emph{Narrative} representations, values are reported as $CGM_{derived}$ / $CGM_{derived}^{context}$. \emph{Raw} corresponds to the $CGM_{only}$ configuration. Bold values indicate the best input configuration for each LLM, prediction task, and horizon. A dash denotes an undefined MCC due to a constant prediction output.}
\label{tab:results_fs}
\setlength{\tabcolsep}{6pt}

\begin{tabular}{lllcccc}
\hline
\textbf{Task} &
\textbf{Horizon} &
\textbf{Representation} &
\textbf{Gemma 2} &
\textbf{Llama 3.1} &
\textbf{Mistral} &
\textbf{Qwen 2.5} \\
\hline

\multirow{9}{*}{Hyperglycemia}
& \multirow{3}{*}{30 min}
& Raw
& \textbf{0.399} & 0.291 & \textbf{0.441} & -- \\
&
& Structured
& 0.370 / 0.345
& 0.077 / 0.099
& 0.395 / 0.319
& 0.145 / \textbf{0.487} \\
&
& Narrative
& 0.265 / 0.273
& 0.324 / \textbf{0.328}
& 0.257 / 0.263
& 0.361 / 0.312 \\
\cline{2-7}

& \multirow{3}{*}{60 min}
& Raw
& \textbf{0.415} & 0.244 & \textbf{0.360} & 0.139 \\
&
& Structured
& 0.193 / 0.235
& 0.111 / 0.109
& 0.227 / 0.245
& 0.045 / \textbf{0.263} \\
&
& Narrative
& 0.190 / 0.185
& \textbf{0.258} / 0.245
& 0.238 / 0.200
& 0.116 / 0.221 \\
\cline{2-7}

& \multirow{3}{*}{90 min}
& Raw
& \textbf{0.388} & \textbf{0.235} & \textbf{0.280} & 0.222 \\
&
& Structured
& 0.285 / 0.302
& 0.091 / 0.090
& 0.273 / 0.171
& 0.324 / \textbf{0.353} \\
&
& Narrative
& 0.145 / 0.144
& 0.190 / 0.198
& 0.170 / 0.187
& 0.178 / 0.244 \\
\hline

\multirow{9}{*}{Hypoglycemia}
& \multirow{3}{*}{30 min}
& Raw
& \textbf{0.360} & \textbf{0.419} & 0.165 & 0.296 \\
&
& Structured
& 0.165 / 0.176
& 0.289 / 0.089
& \textbf{0.453} / 0.349
& 0.305 / \textbf{0.568} \\
&
& Narrative
& 0.149 / 0.154
& 0.111 / 0.118
& 0.170 / 0.138
& 0.338 / 0.340 \\
\cline{2-7}

& \multirow{3}{*}{60 min}
& Raw
& \textbf{0.319} & \textbf{0.437} & 0.345 & 0.312 \\
&
& Structured
& 0.253 / 0.240
& 0.061 / 0.212
& 0.390 / 0.375
& 0.302 / \textbf{0.417} \\
&
& Narrative
& 0.201 / 0.199
& 0.222 / 0.150
& \textbf{0.445} / 0.277
& 0.314 / 0.347 \\
\cline{2-7}

& \multirow{3}{*}{90 min}
& Raw
& \textbf{0.283} & 0.113 & 0.220 & 0.359 \\
&
& Structured
& 0.230 / 0.189
& \textbf{0.219} / 0.087
& 0.397 / \textbf{0.499}
& 0.113 / 0.286 \\
&
& Narrative
& 0.272 / 0.271
& \textbf{0.219} / 0.154
& 0.339 / 0.276
& \textbf{0.403} / 0.402 \\
\hline
\end{tabular}
\end{table*}

Tables~\ref{tab:results_supervised}-\ref{tab:results_fs} provide a comprehensive overview of the direct-classification results.
Table~\ref{tab:results_supervised} reports the MCC achieved by the supervised models across prediction tasks, horizons, and information settings, whereas Tables~\ref{tab:results_zs} and ~\ref{tab:results_fs} report the corresponding results for the general purpose LLMs under ZS and FS inference, respectively, considering the different physiological information representations and information settings.

Based on these complete results, the following subsections provide targeted analyses for the three research questions. We first compare the predictive performance of general purpose LLMs with conventional supervised models (RQ1), then examine how physiological information representation and availability affect prompt-based prediction (RQ2), and finally investigate the regression-based comparison between prompt-based inference and the time series-adapted Gluco-LLM approach (RQ3).

\subsection{RQ1: General purpose LLMs vs.\ supervised learning}
\label{subsec:rq1_results}

RQ1 investigates whether prompt-based general purpose LLMs can achieve predictive performance comparable to conventional supervised models for postprandial glycemic-event prediction. Building on the complete results
reported in Tables~\ref{tab:results_supervised}-\ref{tab:results_fs},
while Table~\ref{tab:rq1_summary} provides a descriptive best-case comparison by reporting, for each prediction task and horizon, the highest test MCC observed among the evaluated supervised, ZS LLM, and FS LLM configurations. For each approach, the best result is selected across all applicable models and information configurations.

\begin{table*}[t]
\centering
\small
\caption{Best observed test MCC among the evaluated supervised, zero-shot LLM, and few-shot LLM configurations for each prediction task and horizon. The best supervised result is selected across models and information settings, whereas the best LLM results are selected across models, representations, and information settings. For stochastic supervised models, values are reported as mean (standard deviation) over 10 runs. Bold values indicate the best overall MCC for each task and prediction horizon.}
\label{tab:rq1_summary}
\setlength{\tabcolsep}{3pt}
\begin{tabular}{lllcclcl}
\hline
\textbf{Task} &
\textbf{Horizon} &
\textbf{Best supervised} &
\textbf{MCC} &
\textbf{Best zero-shot LLM} &
\textbf{MCC} &
\textbf{Best few-shot LLM} &
\textbf{MCC} \\
\hline
\multirow{3}{*}{Hyperglycemia}
& 30 min
& LogReg ($CGM_{derived}$)
& \textbf{0.570}
& Qwen (Structured/$CGM_{derived}^{context}$)
& 0.485
& Qwen (Structured/$CGM_{derived}^{context}$)
& 0.487 \\
& 60 min
& LogReg ($CGM_{derived}^{context}$)
& \textbf{0.523}
& Qwen (Structured/$CGM_{derived}^{context}$)
& 0.407
& Gemma (Raw/$CGM_{only}$)
& 0.415 \\
& 90 min
& XGBoost ($CGM_{derived}^{context}$)
& \textbf{0.455 (0.004)}
& Qwen (Structured/$CGM_{derived}^{context}$)
& 0.375
& Gemma (Raw/$CGM_{only}$)
& 0.388 \\
\hline
\multirow{3}{*}{Hypoglycemia}
& 30 min
& XGBoost ($CGM_{derived}$)
& 0.383 (0.003)
& Gemma (Structured/$CGM_{derived}^{context}$)
& 0.507
& Qwen (Structured/$CGM_{derived}^{context}$)
& \textbf{0.568} \\
& 60 min
& LogReg ($CGM_{derived}$)
& 0.345
& Gemma (Structured/$CGM_{derived}$)
& \textbf{0.456}
& Mistral (Narrative/$CGM_{derived}$)
& 0.445 \\
& 90 min
& XGBoost ($CGM_{derived}$)
& 0.366 (0.002)
& Qwen (Raw/CGM)
& 0.452
& Mistral (Structured/$CGM_{derived}^{context}$)
& \textbf{0.499} \\
\hline
\end{tabular}
\end{table*}

The comparison reveals a clear task-dependent behavior. For hyperglycemia, the best supervised configuration achieves the highest MCC at all three prediction horizons. Although the gap between the best supervised and FS configurations progressively narrows as the prediction horizon increases, the best supervised configuration retains the highest performance at every horizon.

A substantially different pattern emerges for hypoglycemia. In this task, the best prompt-based LLM configuration achieves a higher observed MCC than the best supervised configuration at every prediction horizon, including under ZS inference. The advantage of prompt-based inference is observed across all prediction horizons and is not restricted to the inclusion of in-context demonstrations. One possible contributing factor is the severe class imbalance characterizing the hypoglycemia task, for which the FS setting may partially mitigate the effect of the imbalance by explicitly exposing the model to labeled examples from both classes. This interpretation, however, cannot fully explain the observed advantage, since competitive performance is also obtained under ZS inference. More generally, these results suggest that prompt-based LLMs may be comparatively less affected by the scarcity of positive hypoglycemic examples than conventional supervised models, although further investigation would be required to isolate the mechanisms underlying this behavior.


\subsection{RQ2: Influence of physiological information representation and availability}
\label{subsec:rq2_results}

RQ2 investigates how the representation and the amount of physiological information made available to the model affect prompt-based prediction. Unlike RQ1, which focuses on the relative predictive performance of the considered paradigms, this analysis examines the behavior of the general purpose LLMs across the different input configurations.

The zero-shot results (Table~\ref{tab:results_zs}) do not reveal a general preference for the original CGM trajectory. Indeed, when \emph{Raw} is compared with \emph{Structured} and \emph{Narrative} under $CGM_{derived}$ (thereby keeping the underlying physiological source limited to CGM), \emph{Structured} achieves the highest MCC in the greatest part of the cases in hyperglycemia, while in hypoglycemia the results are more heterogeneous across models and horizons.

Under FS inference (Table~\ref{tab:results_fs}), the relative advantage shifts toward the \emph{Raw} representation, corresponding to the use of $CGM_{only}$ information, which achieves the highest MCC in most model--horizon configurations. Nevertheless, representations based on the $CGM_{derived}$ and $CGM_{derived}^{context}$ information settings yield the best overall few-shot performance for hyperglycemia at 30 minutes and for hypoglycemia at all three prediction horizons. This indicates that \emph{Structured} or \emph{Narrative} representations can still produce the strongest overall result when considering the best-performing model for each task and horizon.

A second aspect of RQ2 concerns whether providing additional physiological information improves prediction. The comparison between $CGM_{derived}$ and $CGM_{derived}^{context}$ does not reveal a systematic advantage for either setting. Under ZS inference, the $CGM_{derived}^{context}$ setting is more frequently preferred for \emph{Structured} inputs, whereas the $CGM_{derived}$ setting clearly performs better for \emph{Narrative} inputs. 

Consequently, the inclusion of insulin, carbohydrate, meal, basal-rate, and exercise information does not consistently improve the predictive performance of the evaluated LLMs. Importantly, this result should not be interpreted as evidence that these variables are predictively irrelevant. Rather, it indicates that the models do not systematically benefit from their inclusion when the additional information is provided through the considered prompt representations.


\subsection{RQ3: Prompt-based inference vs.\ time-series-adapted LLMs}
\label{subsec:rq3_results}

RQ3 investigates whether prompt-based inference with general purpose LLMs can approach the performance of a LLM explicitly adapted to physiological time series forecasting. The comparison is conducted under the regression formulation, using the Cui et al. model as a common supervised reference.
Tab.~\ref{tab:rq3_comparison} summarizes the results in terms of MCC and RMSE, the two metrics that are directly available across all three approaches.

\begin{table*}[t]
\centering
\small
\caption{Regression-based comparison among the Cui et al.\ model, prompt-based general-purpose LLMs under zero-shot and few-shot inference, and Gluco-LLM. For the prompt-based approaches, the best observed test result across the four evaluated LLMs is reported separately for zero-shot and few-shot inference. Cui and Gluco-LLM results are reported as mean (standard deviation) over 10 seeds. Bold values indicate the best overall result for each task, prediction horizon, and metric. Lower RMSE is better.}
\label{tab:rq3_comparison}

\begin{tabular}{llcccc|cccc}
\hline
& &
\multicolumn{4}{c|}{\textbf{MCC}} &
\multicolumn{4}{c}{\textbf{RMSE (mg/dL)}} \\
\textbf{Task} &
\textbf{Horizon} &
\textbf{Cui} &
\textbf{ZS LLM} &
\textbf{FS LLM} &
\textbf{Gluco} &
\textbf{Cui} &
\textbf{ZS LLM} &
\textbf{FS LLM} &
\textbf{Gluco} \\
\hline

Hyperglycemia
& 30 min
& \textbf{0.618 (0.012)}
& 0.456
& 0.560
& 0.577 (0.007)
& \textbf{18.45 (0.21)}
& 23.64
& 21.53
& 20.79 (0.26) \\

& 60 min
& \textbf{0.492 (0.008)}
& 0.343
& 0.353
& 0.471 (0.014)
& \textbf{30.53 (0.22)}
& 40.77
& 37.67
& 35.38 (0.22) \\

& 90 min
& \textbf{0.436 (0.013)}
& 0.234
& 0.231
& 0.379 (0.015)
& \textbf{37.64 (0.53)}
& 52.72
& 46.76
& 46.80 (0.33) \\

\hline

Hypoglycemia
& 30 min
& \textbf{0.470 (0.028)}
& 0.265
& 0.409
& 0.460 (0.044)
& \textbf{13.75 (0.23)}
& 17.75
& 15.45
& 15.08 (0.45) \\

& 60 min
& 0.380 (0.043)
& 0.186
& 0.240
& \textbf{0.397 (0.030)}
& \textbf{23.94 (0.29)}
& 27.44
& 27.62
& 26.12 (0.31) \\

& 90 min
& 0.299 (0.056)
& 0.222
& \textbf{0.378}
& 0.285 (0.040)
& \textbf{31.51 (0.46)}
& 35.50
& 34.11
& 35.26 (0.26) \\

\hline
\end{tabular}
\end{table*}

The comparison with the prompt-based LLMs shows a clear advantage for the task-specific model: Cui et al. model achieves the highest MCC in 4 of the 6 task--horizon combinations and the lowest RMSE in all six. The gap is relatively small for short-horizon hyperglycemia, but increases substantially at longer horizons. At 90 minutes, for example, the corresponding MCC values are 0.234 and 0.436, respectively. The only case in which a prompt-based LLM exceeds the Cui et al. baseline in MCC is hypoglycemia at 90 minutes; however, its continuous glucose prediction remains less accurate in terms of RMSE.

Considerng instead Gluco-LLM, its MCC is closer to that of Cui et al. in most experimental conditions, particularly for hypoglycemia. However, this difference is small relative to the variability observed across different runs.
A more consistent difference emerges for continuous glucose prediction. Cui et al. achieves a lower RMSE than Gluco-LLM in every task-horizon combination. Thus, the competitive event-detection performance occasionally observed for Gluco-LLM does not translate into more accurate prediction of the underlying glucose extreme.

Within the regression-based comparison reported in Tab.~\ref{tab:rq3_comparison}, Gluco-LLM remains closer to the Cui et al. model than the best observed prompt-based LLM configuration in most conditions, suggesting that explicit adaptation to physiological time series may enable a more effective use of a pretrained language-model backbone than prompt-based inference without task-specific parameter adaptation. This advantage is not universal, however: prompt-based inference obtains the highest LLM MCC for hypoglycemia at 90 minutes and marginally lower RMSE in two of the six conditions.

Overall, RQ3 shows that none of the investigated LLM approaches consistently achieves the performance of the task-specific model. Indeed, the Cui et al. model currently remains the strongest and most consistent approach for the postprandial regression tasks considered in this study.
\section{Discussion}
\label{sec:discussion}
In this work, we have provided a detailed investigation of the capabilities and limitations of language model-based approaches for postprandial glycemic-event prediction. Across the three research questions, the results reported in Section~\ref{sec:results} show that general purpose LLMs can produce non-trivial predictions without task-specific parameter optimization. However, their performance strongly depends on the model, physiological representation, prompting strategy, target event, and prediction horizon.

\paragraph{General purpose LLMs can be competitive in some settings} 

The results for RQ1 reported in Tables~\ref{tab:results_supervised}-~\ref{tab:rq1_summary} show that prompt-based LLMs cannot yet be regarded as a general replacement for conventional supervised models. For hyperglycemia, the best observed supervised configuration achieves the highest MCC at every prediction horizon. Conversely, for hypoglycemia, the best observed prompt-based configuration outperforms the best observed supervised result at all three horizons, including under ZS inference. 

Furthermore, the effectiveness of prompt-based inference is strongly configuration-dependent. Performance varies across LLM families, prompting strategies, physiological representations, target events, and prediction horizons. The best observed FS configuration outperforms the best observed ZS configuration in the greatest part of  task--horizon combinations. Similarly, no representation is uniformly optimal: \emph{Structured} produces the best overall result in most task--prompting--horizon combinations, whereas \emph{Raw} and \emph{Narrative} remain preferable in specific settings.

This task-dependent behavior indicates that general purpose LLMs can extract predictive information from short glucose histories and derived physiological descriptors, even though numerical time series modeling is not their native training objective.
However, LLMs cannot yet be regarded as a consistently superior replacement for models trained specifically for glycemic-event prediction.
Therefore, the comparison should not be framed solely in terms of which paradigm ``wins''. Rather, it reveals a trade-off between the overall reliability and consistency of supervised models and the training-free adaptability, accessibility, and representation flexibility offered by prompt-based LLM inference.

\paragraph{How physiological information is represented could matter at least as much as the kind of information provided} 

The results do not identify a representation of the information that is uniformly preferable across all experimental conditions. As reported in Section~\ref{sec:results}, descriptor-based representations can produce the strongest overall configurations, even though preserving the original CGM sequence remains advantageous for several individual models.

The comparison between $CGM_{derived}$ and $CGM_{derived}^{context}$ further shows that increasing the amount of physiologically relevant information is not sufficient to ensure better prompt-based inference. Consequently, the effect of insulin, carbohydrate, basal-rate, meal, and exercise information depends on how these variables are represented and integrated into the prompt rather than simply on their availability.

FS prompting also exhibits a configuration-dependent behavior. Although it improves the best observed result in several cases, it increases MCC in only a limited number of cases of the matched comparisons. In-context examples should therefore not be regarded as uniformly beneficial; their effectiveness depends on the model, representation, information setting, task, and prediction horizon.

Taken together, these findings show that representation design is not a secondary implementation choice when LLMs are applied to physiological time series. Neither preserving the original signal nor transforming it into structured or narrative descriptors provides a universal advantage. Instead, the most effective interface depends on the interaction among the input representation, the available physiological information, and the specific LLM.

\paragraph{Explicit time-series adaptation narrows, but does not close, the gap}

RQ3 examines whether the limitations observed for prompt-based inference can be mitigated by explicitly adapting a pretrained language-model backbone to physiological time-series forecasting. Gluco-LLM generally provides a stronger competitor to the Cui et al. model than general purpose models, particularly for hypoglycemia, indicating that learned time series interfaces can exploit a language model backbone more effectively than textual prompting alone.

Nevertheless, the Cui et al. model remains the most reliable model in the regression experiments. It achieves the lowest RMSE in every considered task--horizon combination and the strongest MCC in almost all comparisons with Gluco-LLM. The isolated MCC advantage obtained by Gluco-LLM for hypoglycemia at 60 minutes is small relative to the variability across runs and therefore does not indicate a systematic superiority of the adapted LLM. This is consistent with the observation that event detection after thresholding does not necessarily imply an equally accurate estimate of the underlying glucose value.

The result is also informative with respect to the applicability of language-model-based time series architectures. Gluco-LLM was originally developed for continuous glucose forecasting, whereas the present study considers a distinct prediction setting based on short, meal-centered observation windows and future glucose-extreme estimation. This difference in task formulation may partly explain why its behavior in the present postprandial setting differs from that observed in conventional glucose forecasting \cite{li2025llm}.


\section{Limitations and future work}
\label{sec:limitations}
The findings of this study should be interpreted in light of several limitations, which also identify relevant directions for future research.

First, the experiments rely exclusively on the OhioT1DM dataset. Although this dataset provides a well-established benchmark for glucose prediction and enables patient-specific evaluation under a controlled chronological protocol, conclusions drawn from a single dataset cannot establish the generality of the observed behavior. Differences in population characteristics, sensing devices, treatment strategies, and data-collection protocols may affect both supervised models and LLM-based approaches. Future work should therefore reproduce the proposed evaluation framework on additional CGM datasets and, where possible, assess cross-dataset robustness. A related limitation concerns the training data of the adopted LLMs.
Because the complete pretraining corpora are not publicly available, it is not possible to exclude with certainty that the evaluated models were exposed to OhioT1DM data, derived material, or closely related distributions during pretraining \cite{apicella2025don}. Such contamination cannot be quantified within the present study. Future work should investigate strategies for evaluating pretraining contamination and memorization in biomedical time-series benchmarks, particularly when general purpose foundation models are compared with models trained exclusively on the experimental training partition.

Second, the prompt-based experiments consider open-weight language models in the 7-9B parameter range. This choice favors local execution, reproducibility, and experimental control, but it does not characterize the behavior of larger open-weight models or proprietary frontier models. The conclusions should therefore be restricted to the class of models investigated here. Evaluating a broader range of model scales and architectures would help determine whether the observed limitations are intrinsic to prompt-based physiological reasoning or partly attributable to model capacity.

Third, the variability of prompt-based inference is less extensively characterized than that of the trained models. Supervised stochastic models, the Cui et al. model, and Gluco-LLM are evaluated over multiple random runs, where applicable, to account for variability introduced by stochastic model training and initialization. In contrast, prompt-based LLM configurations do not involve task-specific training or parameter updates; therefore, each configuration is evaluated through a single inference run under a fixed prompting setup. Consequently, the reported differences do not quantify sensitivity to alternative demonstration sets or other sources of inference-time variability.
Future studies should repeat prompt-based experiments across multiple decoding configurations, including different temperature settings, and report the corresponding variability and uncertainty estimates.

Finally, the investigated representations cover only a limited subset of the ways in which multimodal physiological information can be presented to an LLM. In particular, the lack of a systematic benefit from the additional contextual variables should not be interpreted as evidence that insulin, carbohydrate, meal, or physical-activity information is intrinsically uninformative. Rather, this finding is specific to the way such information is represented and incorporated in the present study. Future work could investigate representations that retain the temporal evolution of insulin, meal, carbohydrate, and physical-activity information, rather than summarizing these variables into aggregate descriptors.

\section{Conclusion}
\label{sec:conclusion}

This study investigated the use of language models for postprandial glycemic-event prediction in individuals with T1DM, comparing three complementary paradigms: conventional supervised ML, prompt-based inference with general-purpose LLMs, and time series adaptation of a pretrained LLM. The analysis considered both direct event classification and regression of future glucose extrema under a common patient-specific and temporally causal evaluation framework.

With respect to RQ1, the comparison did not establish a uniform ranking between supervised ML and LLM-based inference, indicating that general purpose LLMs are not intrinsically unsuitable for glucose prediction, although their effectiveness is strongly dependent on the target event and experimental configuration. From the observed results, few-shot prompting should be regarded as a configuration-dependent adaptation mechanism rather than as a uniformly beneficial strategy.

RQ2 showed that the representation of physiological information has a substantial effect on prompt-based prediction, but did not identify a universally optimal representation. When the best result was selected across models, \emph{Structured} inputs achieved the highest MCC in most task--prompting--horizon combinations. Conversely, \emph{Raw} was more frequently preferred across matched model configurations under FS inference, while \emph{Narrative} became competitive in specific hypoglycemia settings. The effectiveness of a representation therefore depends on the interaction among the LLM family, prompting strategy, target event, and prediction horizon.

Similarly, expanding the input from $CGM_{derived}$ to $CGM_{derived}^{context}$ did not produce a systematic improvement. The additional insulin, carbohydrate, meal, basal-rate, and exercise information was beneficial in some configurations but not in others, and its effect varied with the adopted representation. This does not imply that these variables are physiologically or predictively irrelevant. Rather, it suggests that making additional information available is not sufficient unless the model can effectively integrate it through the chosen input interface. This aspect is particularly relevant when considering the potential use of LLMs for hypoglycemic or hyperglycemic event prediction in patient-oriented applications, where their ability to process heterogeneous information through a natural language interface could help reduce some of the theoretical and practical barriers associated with standard ML approaches. In this context, however, the effective representation and integration of clinically relevant information remains a key requirement for translating such flexibility into reliable predictive performance.

Finally, RQ3 showed that explicitly adapting a language model backbone to physiological time series narrows the gap with task-specific forecasting models but does not eliminate it. In particular, Gluco-LLM was generally more competitive with the Cui et al. model than prompt-only inference, particularly for hypoglycemia. Nevertheless, the recurrent baseline achieved a lower RMSE in every evaluated condition and the highest MCC in most comparisons. Thus, explicit time series adaptation appears more effective than prompt-only inference for exploiting a pretrained language model backbone, but the task-specific recurrent model remains the strongest and most consistent approach for the considered regression task.

Taken together, these results suggest that a central challenge for LLM-based glucose prediction lies in effectively representing and modeling patient-specific temporal dynamics. Specialized models trained directly on physiological data remain the most reliable choice in several experimental conditions, particularly for hyperglycemia classification and continuous glucose prediction. However, their advantage is not universal: appropriately selected general purpose LLM configurations achieve competitive or superior event-detection performance for hypoglycemia without task-specific parameter training. The resulting picture is therefore not one of uniform superiority, but of a trade-off between the consistency of task-specific models and the training-free adaptability and representational flexibility of prompt-based inference.

Future work should extend the evaluation to additional glucose-monitoring datasets, larger and more diverse language models, repeated prompt-based inference protocols, and adaptation strategies capable of preserving the temporal relationships among heterogeneous physiological and therapeutic signals. Dedicated ablation studies should also examine the effect of the number, selection, and class distribution of in-context demonstrations, particularly for severely imbalanced prediction tasks.

\section*{Acknowledgement}
The authors would like to thank Prof. Giovanni Annuzzi for his valuable clinical insights and constructive discussions, which contributed to the framing and interpretation of this work.

This work was partially funded by the PNRR MUR project PE0000013-FAIR (CUP: E63C25000630006).



\bibliographystyle{elsarticle-num}
\bibliography{__BIB,__BIB_SELF}

\end{document}